\documentclass[acmsmall,screen]{acmart}

\def\ie{\emph{i.e.}} 

\usepackage{amsmath,amsfonts,bm}

\def\eqref#1{equation~\ref{#1}}

\def\1{\bm{1}}

\def\va{{\bm{a}}}

\def\vf{{\bm{f}}}

\def\vh{{\bm{h}}}

\def\vx{{\bm{x}}}

\def\vz{{\bm{z}}}

\def\mA{{\bm{A}}}

\def\mE{{\bm{E}}}
\def\mF{{\bm{F}}}
\def\mG{{\bm{G}}}
\def\mH{{\bm{H}}}

\def\mW{{\bm{W}}}
\def\mX{{\bm{X}}}

\def\mZ{{\bm{Z}}}

\DeclareMathAlphabet{\mathsfit}{\encodingdefault}{\sfdefault}{m}{sl}
\SetMathAlphabet{\mathsfit}{bold}{\encodingdefault}{\sfdefault}{bx}{n}
\newcommand{\tens}[1]{\bm{\mathsfit{#1}}}

\def\tG{{\tens{G}}}
\def\tH{{\tens{H}}}

\def\tO{{\tens{O}}}

\def\gG{{\mathcal{G}}}

\def\gS{{\mathcal{S}}}
\def\gT{{\mathcal{T}}}

\newcommand{\R}{\mathbb{R}}

\usepackage{multirow}
\usepackage{booktabs}
\usepackage{makecell}
\usepackage{color}

\newcolumntype{L}[1]{>{\raggedright\let\newline\\\arraybackslash\hspace{0pt}}m{#1}}
\newcolumntype{C}[1]{>{\centering\let\newline\\\arraybackslash\hspace{0pt}}m{#1}}
\newcolumntype{R}[1]{>{\raggedleft\let\newline\\\arraybackslash\hspace{0pt}}m{#1}}

\AtBeginDocument{%
  }

\setcopyright{acmcopyright}
\copyrightyear{2023}
\acmYear{2023}
\acmDOI{XXXXXXX.XXXXXXX}

\acmJournal{TOMM}
\acmVolume{37}
\acmNumber{4}
\acmArticle{111}
\acmMonth{8}

\begin{document}

\title{Building Category Graphs Representation with Spatial and Temporal Attention for Visual Navigation}

\author{Xiaobo Hu}
\email{xiaobohu@bjtu.edu.cn}
\author{Youfang Lin}
\email{yflin@bjtu.edu.cn}
\affiliation{%
  \institution{Beijing Jiaotong University}
  \country{China}
}
\author{HeHe Fan}
\email{crane.h.fan@gmail.com}
\affiliation{%
  \institution{Zhejiang University}
  \country{China}
}
\author{Shuo Wang}
\email{shuo.wang@bjtu.edu.cn}
\author{Zhihao Wu}
\email{zhwu@bjtu.edu.cn}
\author{Kai Lv}
\email{lvkai@bjtu.edu.cn}
\affiliation{%
  \institution{Beijing Jiaotong University}
  \country{China}
}
\renewcommand{\shortauthors}{X.Hu et al.}

\begin{abstract}
Given an object of interest, visual navigation aims to reach the object's location based on a sequence of partial observations. 
To this end, an agent needs to 1) learn a piece of certain knowledge about the relations of object categories in the world during training and 2) look for the target object based on the pre-learned object category relations and its moving trajectory in the current unseen environment. 
In this paper, we propose a Category Relation Graph (CRG) to learn the knowledge of object category layout relations and a Temporal-Spatial-Region (TSR) attention architecture to perceive the long-term spatial-temporal dependencies of objects helping the navigation.
We learn prior knowledge of object layout, establishing a category relationship graph to deduce the positions of specific objects.
Subsequently, we introduced TSR to capture the relationships of objects in temporal, spatial, and regions within the observation trajectories.
Specifically, we propose a Temporal attention module (T) to model the temporal structure of the observation sequence, which implicitly encodes the historical moving or trajectory information. 
Then, a Spatial attention module (S) is used to uncover the spatial context of the current observation objects based on the category relation graph and past observations. 
Last, a Region attention module (R) shifts the attention to the target-relevant region. 
Based on the visual representation extracted by our method, the agent can better perceive the environment and easily learn superior navigation policy. 
Experiments on AI2-THOR demonstrate our CRG-TSR method significantly outperforms existing methods regarding both effectiveness and efficiency. 
The code has been included in the supplementary material and will be publicly available.
\end{abstract}




\maketitle

\section{Introduction}
In visual navigation~\cite{meng1993mobile,DBLP:conf/icra/BloschWSS10,cummins2007probabilistic,dissanayake2001solution}, an agent aims to find the target object based on monocular visual observations. 
To move toward the goal, the agent needs to first correctly perceive the environment and then take effective actions. 
Although a number of reinforcement learning approaches \cite{ng2003autonomous,kohl2004policy,mnih2015human,peters2008reinforcement,wang2023skill} have been proposed to learn effective policies for visual navigation, the prerequisite, \ie, precisely capturing the relations of objects in the scene and understanding the entire environment, still remains challenging, especially based on egocentric observations. 
This paper focuses on the challenge of environment layout information perception.
We aim to extract prior knowledge about object layout relationships from the training environment and utilize historical trajectory observations to better perceive the navigation scene layout.

\begin{figure}
  \centering
  \includegraphics[width=\linewidth]{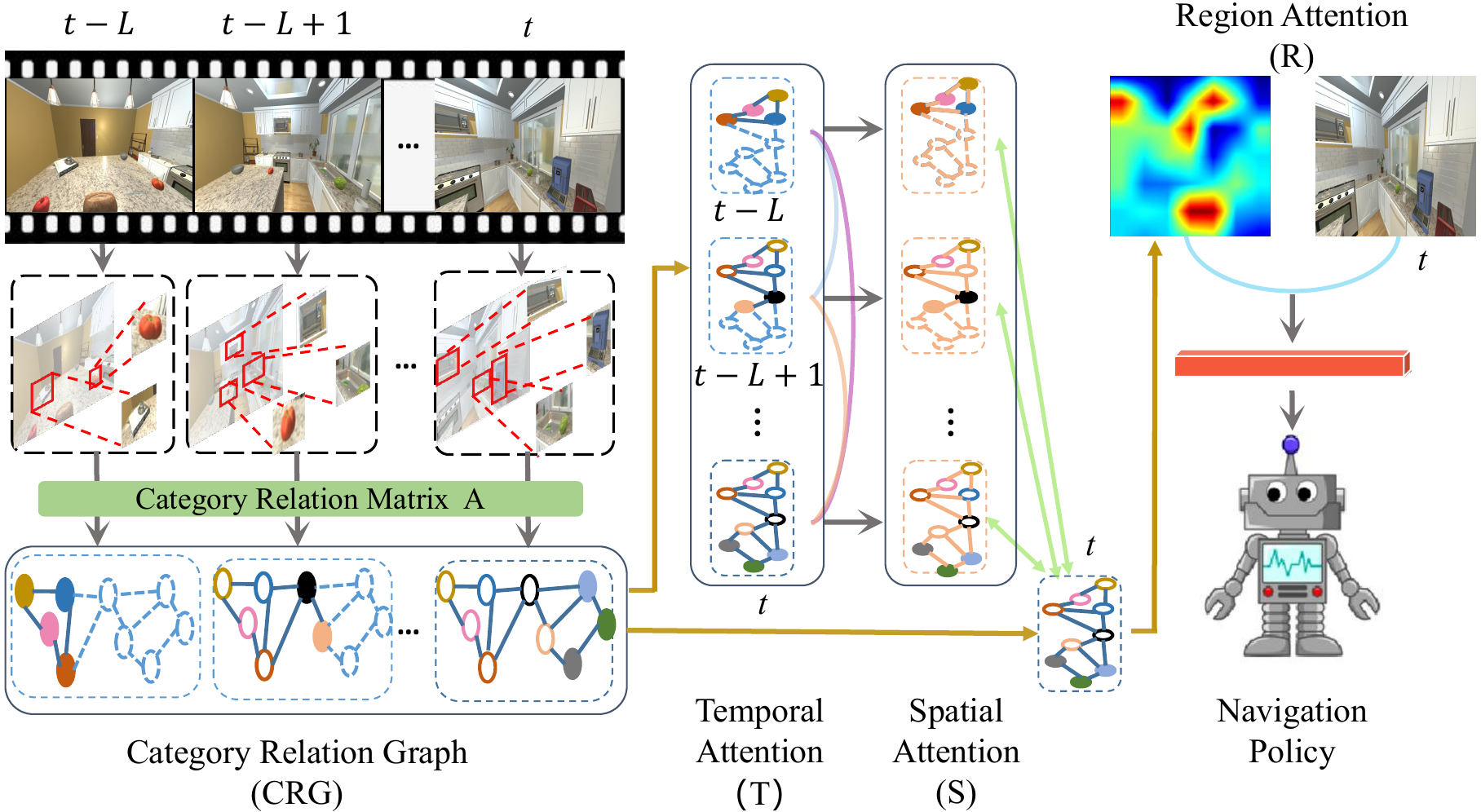}
  \caption{\textbf{Illustration of our visual navigation method.}
Our method learns a Category Relation Graph (CRG) to embed the knowledge of object category relations. 
Then, we propose a Temporal attention module (T) to model the temporal structure of the observation sequence and a Spatial attention module (S) to uncover the spatial context of the current observation.
Finally, a Region attention module (R) shifts the attention to the target-relevant region and obtains the final visual representation for the navigation policy learning.
  }
  \label{intro}
\end{figure}

To effectively locate navigation target objects, the agent needs to learn prior knowledge about the layout relationships of object categories in the scene.
For example, when looking for a microwave in an unfamiliar room layout, the agent should know that microwaves are usually in kitchens alongside sinks and stove-burners. 
Furthermore, when the agent finds the stove-burner, it should be able to infer that the microwave is just around here. 
This prior knowledge can provide an important cue for visual navigation. 
Without prior knowledge of object layout, the agent has to collect the layout information of the unseen environment through massive trials and errors \cite{DBLP:conf/icra/LiangCS21}, thus leading to inferior efficiency and even failed navigation.
To learn the prior information about the object layout relationship, we propose a Category Relation Graph (CRG) to embed the knowledge of object category relations, as shown in Figure~\ref{intro}. 
Although prior works~\cite{mayo2021visual,du2020learning} have utilized graphs in visual navigation, they usually build relationship graphs based on single-frame observations~\cite{du2020learning}, roughly established regional relationships~\cite{zhang2021hierarchical}, or involve extensive exploration steps to model scene maps \cite{DBLP:conf/icra/LiangCS21}. 
In contrast, our approach utilizes trajectory observations to continually update the global object relationship graph, enabling adaptive learning of the empirical knowledge about object layout relationships.
During testing, we utilize the acquired category relation graph to deduce the locations of objects, thus guiding the agent's perception of the unfamiliar environment.


To enhance the agent's ability to perceive the scene layout and perform navigation tasks more effectively, the agent need not only possess prior knowledge of the scene layout but also retain the experiences from past observations to prevent memory loss. 
The navigation trajectory observation contains valuable scene layout knowledge. 
Previous works~\cite{liu2022tcgl,liu2022cross,guo2021learning} typically store historical information and extract attention from observations in a simplistic manner.
For example, prior works~\cite{pathak2018zero,fang2019scene} neglect the guidance of object-related information.
For OMT~\cite{DBLP:conf/icra/FukushimaOKSY22}, although it introduces object guidance, it merely stores it without delving into in-depth modeling for the object.
However, we aim to model trajectory history information in detail to enhance the agent’s perception of the layout relationships in the scene.
We introduce a Temporal-Spatial-Region attention architecture (TSR) designed to capture long-term spatial-temporal dependencies within observation objects, as shown in Figure~\ref{intro}.
Specifically, temporal attention aims to capture the historical information from the graph sequence with the extended temporal receptive field, enabling the extraction of layout information from the temporal dimension.
After that, spatial attention uncovers the spatial context of the current observation category nodes and collects the most necessary information from historical observation. 

Last, a region attention module shifts the attention to the target-relevant region of the observation, which can be used to guide navigation actions.
As a result, we have obtained information-rich visual representations that assist the agent in accurately perceiving the environment, leading to high-performance visual navigation.

After obtaining the visual representation, we implement the Asynchronous Advantage Actor-Critic (A3C) architecture \cite{DBLP:conf/iclr/BabaeizadehFTCK17} to learn the navigation policy. 
In addition, we utilize supervised learning for the pre-training process of the visual representation to speed up the training convergence. 
We conduct experiments on the artificial simulation environment AI2-THOR \cite{kolve2017ai2}, and the results demonstrate that our method achieves superior navigation performance compared to the previous state-of-the-art methods.
We improve the Success Rate (SR) from 79.6\% to 80.0\% and Success weighted by Path Length (SPL) from 0.449 to 0.457. 
Our contributions are threefold: 
\begin{itemize}
    \item 
    We introduce a Category Relationship Graph (CRG) that utilizes multiple observations to learn inherent relationships between category layouts as prior knowledge.
    \item 
    We imply a Temporal-Spatial-Region attention architecture (TSR) that models object information in trajectory to enhance the long-term layout relationship perception.
    \item We achieve state-of-the-art accuracy on the widely-used AI2-THOR navigation simulator and surpass the existing methods by large margins.
\end{itemize}

\section{Related Work}
\subsection{Rule-based Navigation} 
Visual navigation task is an important problem in embodied artificial intelligence.
Traditional approaches \cite{DBLP:conf/icra/BloschWSS10,cummins2007probabilistic,dissanayake2001solution,thrun1998learning,kidono2002autonomous} usually employ an entire map of an environment and then perform path planning.
For example, Kidono \textit{et al.} propose a navigation policy that relies on the assistance of humans to map the environment~\cite{kidono2002autonomous}.
Oriolo \textit{et al.} present a practical method to automatically build maps online by fusing laser and sonar data to improve navigation performance~\cite{oriolo1995line}.
In addition, other methods usually employ the given map to escape obstacles in the environment rather than for the navigation task~\cite{borenstein1989real,borenstein1991vector}. 
Wei \textit{et al.} present a method to efficiently understand indoor scenes from a single image without training or knowledge of the camera’s internal calibration~\cite{wei2018visual}.
Liang \textit{et al.} utilize the confidence-aware semantic scene completion module to explicitly model the scene and then guide the agent’s navigation actions~\cite{DBLP:conf/icra/LiangCS21}.
Generally, most previous works treat this problem as a geometric problem and utilize additional maps or parameters to assist in planning their path actions. 
However, these methods can not be generalized to unseen or novel environments because these environment maps are not available.

\subsection{Learning-based Navigation}
Recently, deep neural networks have made tremendous achievements, driven by fast simulators \cite{kolve2017ai2,savva2019habitat}, large-scale realistic datasets \cite{DBLP:conf/3dim/ChangDFHNSSZZ17,DBLP:conf/nips/RamakrishnanGWM21,xia2018gibson}, and advanced network architecture \cite{yu2019spatial,yu2019hierarchical}. 
Mirowski \textit{et al.} propose auxiliary methods for prediction and loop closure classification for navigation problems in 3D maze environments~\cite{DBLP:conf/iclr/MirowskiPVSBBDG17}.
Meng \textit{et al.} introduce a reachability estimator that provides the agent with a trackable sequence of target observations \cite{meng2020scaling}.
With the rise of Reinforcement Learning (RL), many methods employ RL to solve the visual navigation problem~\cite{ng2003autonomous,kohl2004policy,mnih2015human,peters2008reinforcement}.
Usually, those methods utilize visual observations as state input and learn the navigation policy. 
For example, Wortsman \textit{et al.} propose a meta-reinforcement learning method to help agents learn to learn in an unseen environment \cite{wortsman2019learning}.
Du \textit{et al.} introduced a Transformer to combine a spatial-enhanced local descriptor and a position-encoded global descriptor, learning the relationships between instances and observation regions \cite{DBLP:conf/iclr/DuY021}.
Dang \textit{et al.} propose a dual adaptive thinking method that flexibly adjusts policy in different navigation stages \cite{dang2023search}.
Xie \textit{et al.} propose a novel implicit obstacle map-driven indoor navigation framework for robust obstacle avoidance \cite{xie2023implicit}.
Ramakrishnan \textit{et al.} realize the decoupling of scene perception and navigation actions,  helping to find the target object \cite{ramakrishnan2022poni}.

Wang \textit{et al.} propose a skill-based hierarchical SHRL method that contains a high-level policy and three low-level skills \cite{wang2023skill}.
Other works introduce more environmental information to improve navigation performance \cite{DBLP:conf/rss/ChenVSXSVS19,DBLP:conf/iclr/SavinovDK18,DBLP:journals/corr/abs-1807-06757,10.1145/3526024,10.1145/3550485}.
However, it remains challenging to effectively perceive the environment for navigation.  

\subsection{Relational Models for Navigation}
Among the learning-based methods, relational models have been applied to visual navigation~\cite{du2020learning,wortsman2019learning,DBLP:conf/iclr/YangWFGM19,DBLP:conf/iclr/PuigSLWLTF021}.  
For example, Yang \textit{et al.} establish an object graph by extracting the relationships among object categories~\cite{DBLP:conf/iclr/YangWFGM19} in Visual Genome~\cite{krishna2017visual}.
This method only employs word embedding to establish the object relationships.
Du \textit{et al.} learn the concurrence relationships of objects from single observations~\cite{du2020learning}.
Dang \textit{et al.} consider the attention bias caused by object visibility and propose a directed object attention graph~\cite{DBLP:conf/mm/DangSWHLC22}.
However, they only focus on the single-frame observation, ignoring the inherent layout relationships in multiple observations.
Other methods consider the relationships among multiple observations in navigation trajectories \cite{pathak2018zero,wu2019bayesian,fang2019scene}.
Pathak \textit{et al.} stack several LSTM modules in a policy network to enhance the temporal memory \cite{pathak2018zero}. 
Fang \textit{et al.} directly embedded each observation into memory and employed a Transformer to perceive the long-term spatial-temporal dependencies~\cite{fang2019scene}.
Fukushima \textit{et al.} consider the impact of long-term memory on navigation and introduce a memory storage module to prevent memory forgetting~\cite{DBLP:conf/icra/FukushimaOKSY22}.
Unlike those methods, our algorithm is methodologically groundbreaking because it simultaneously models the global object category relationship graph and models trajectory history information in detail to enhance the agent's perception of the layout relationships in the scene. 


\section{Proposed Visual Navigation Method}
This section begins by introducing and formulating the visual navigation problem. 
We then describe the construction of a Category Relation Graph (CRG) based on multiple observations. 
Next, we describe the Temporal-Spatial-Region attention architecture (TSR) in detail. 
Finally, we combine the proposed method CRG-TSR and reinforcement learning to tackle the visual navigation problem.

\subsection{Problem Formulation}

In the object visual navigation task, the agent does not have additional sensors or prior knowledge datasets about the environment. 
The agent utilizes the trained scene perception and policy modules to perform navigation actions in the unfamiliar environment and reach the given target object.
The training and testing rooms have the same target object categories set, but they have different object layouts and room styles.
The challenge of the navigation task is to reach the target in an unseen room layout.
In every episode starting step,  the navigation room and the target object $T\in \left \{Television, \cdots, Sink \right \}$ are chosen randomly for the agent.
The initialized state of the agent can be expressed as $s=(x,y,\theta _{r},\theta _{h})$, and the initial observation is $\tens{O}$.
$x$ and $y$ refer to the layout coordinates of the agent, while $\theta _{r}$ and $\theta _{h}$ represent the rotation and horizontal angles of the agent camera.
At step $t$, the agent takes an action based on a learned policy $\pi(\va_t|\tO_t, \vh_{t-1})$, where $\va_t$ denotes the action distribution, observation $\tO_t \in \R^{C\times H \times W}$ is a vector of the visual RGB pixel, and $\vh_{t-1}$ is the previous state of policy. 
Agents can interact with the environment with 6 types of action, \textit{i.e.,} MoveAhead, RotateLeft, RotateRight, LookUp, LookDown, and Done.
After the agent performs an action, it will be in the next state $s'=(x,y,\theta _{r},\theta _{h})$ and get a new image observation $\tens{O}_{t+1}$ of the environment.
In training, the optimization goal is to achieve the maximum value of the cumulative reward $R=\sum_{t=1}^{L_{n}}\gamma^{t}r_t$, where $r_{t}$ denotes the reward given by the environment, $\gamma$ is a discount factor, and $L_{n}$ is the episode length.
The agent receives a negative reward of $-0.01$ for each step and a positive reward of $5$ for the final success step.
The task's success requires the agent to capture and get close to the target object (less than $1.5$ meters).

\subsection{Category Relation Graph}
To efficiently and effectively estimate the location of the target object in an unseen environment, the agent needs prior knowledge about the relations of the object categories in the world. 
Without prior knowledge, the agent has to capture the layout structure of the unseen environment through massive trials and errors, which is computationally expensive. 
Moreover, because trials and errors are not always reliable, they may lead to inferior efficiency and even failed navigation. 
To address this problem, we propose learning a category relation graph to embed object category relations. 
This graph can be used as prior knowledge for other unseen test rooms with the same layout distribution with training.  

\subsubsection{Object Detection and Depth Estimation} 
To build the category relation graph, we first use DETR \cite{carion2020end} as the detector to acquire object locations and appearance features.
Given an observation input $\tO_t$, for each object, the detection results include a bounding box $[x_1,y_1;x_2,y_2]$, a detection confidence value $c$, a semantic label $s$, 
and an extracted appearance feature $\vf \in \R^{1 \times C_{a}}$,
where $(x_1,y_1)$ and $(x_2,y_2)$ are the coordinates of the upper left corner and lower right corner of the object in observation, respectively. 
Then, to estimate the depth value of objects, we employ the pre-trained depth estimation model SARPN~\cite{DBLP:conf/ijcai/ChenCZ19} to generate the depth map of the observation. 
We approximate the object depth value $d$ using the mean value of the depth map within the bounding box $[x_1,y_1;x_2,y_2]$.

\begin{figure}
  \centering
  \includegraphics[width=1\linewidth]{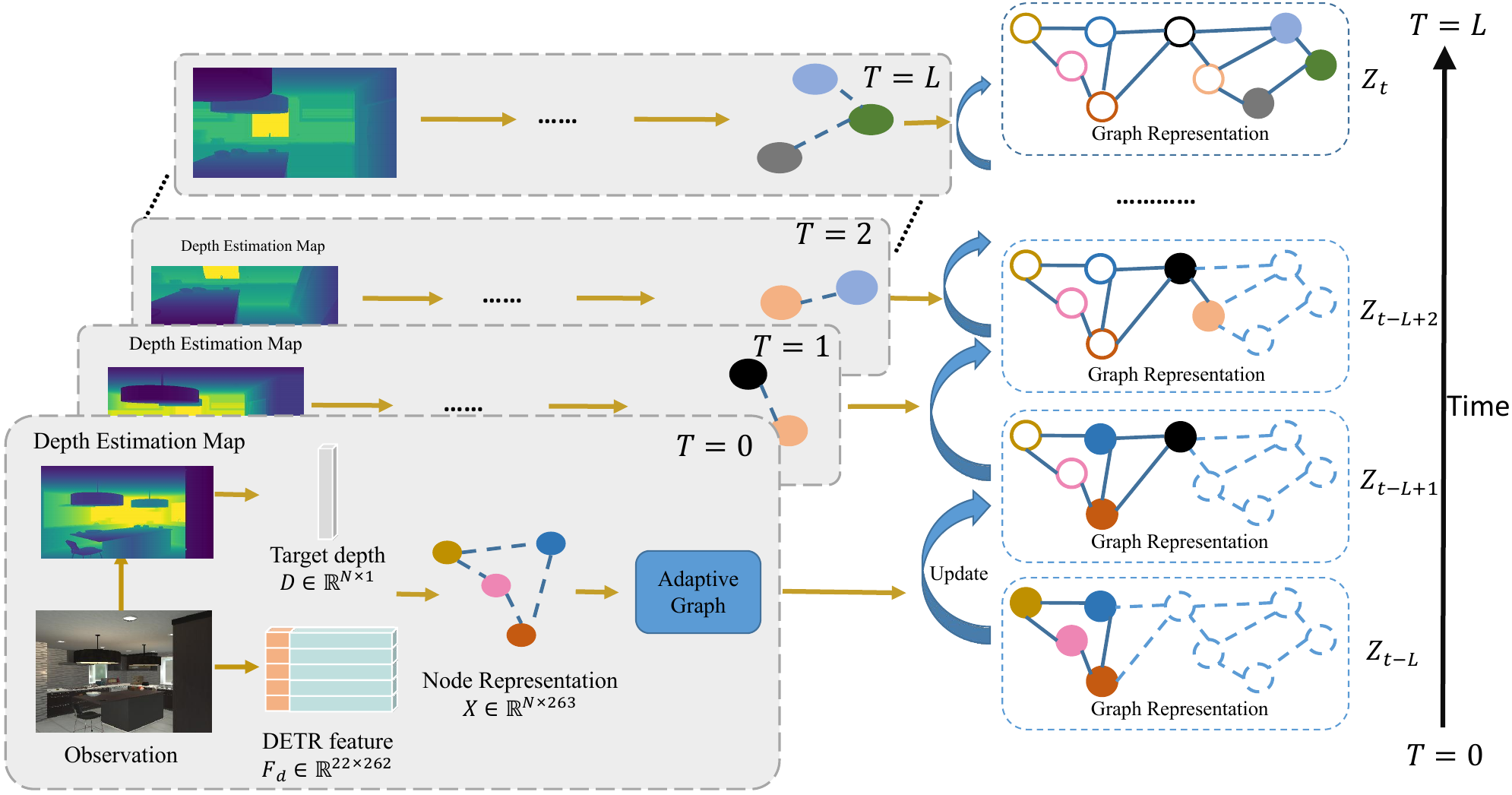}
  \caption{\textbf{Illustration of constructing the category relation graph. }
We integrate a category adjacency matrix in training and utilize it to encode the knowledge of the object into a category relation graph. 
Since the visible object at each step differs, we generate a corresponding graph representation sequence.
Note the graph sequence is utilized for encoding the final visual representation, and the adjacency matrix is updated based on reinforcement learning rewards during training.
  }
  \label{Complete_Graph}
\end{figure}

\subsubsection{Graph Construction} 
Based on the object detection and depth estimation results, we use a graph  $\gG=(\mX, \mA)$ to establish object category relations, as shown in Figure~\ref{Complete_Graph}. 
$\mX \in \R^{N\times C_n}$ represents the graph nodes,
which include the objects' coordinates $[x_1,y_1;x_2,y_2]$, the detection semantic label $s$, the confidence $c$,  the depth value $d$,
and the appearance feature $\mF_{a} \in \mathbb{R}^{ N \times C_{a}}$.
$N$ is the number of object categories in the environment, which is equal to 22 in this paper. 
The adjacency matrix $\mA \in \R^{N\times N}$ represents the relationships among all object categories. 
Note that our graph is category-based. 
When there are multiple instances of the same class, we randomly sample one instance to represent the class in each observation. 
In this way, our category relation matrix can learn all the possible relations with enough training and be utilized to obtain the graph representation.

Following \cite{DBLP:conf/ijcai/WuPLJZ19}, we randomly initiate two node embedding dictionaries with learnable parameters $\mE_1, \mE_2\in \R^{N\times C}$ to adaptively learn the global category adjacency matrix $\mA$, where the $C$ is the dimension of the dictionaries, which can be formulated as follows: 
\begin{equation}
  \mA=\mathrm{Softmax}(\mathrm{ReLU}\big(\mE_1 \mE_2^\top)\big). 
\end{equation}
Then, we employ a graph convolutional network (GCN)~\cite{velickovic2017graph} to obtain the category relation graph representation $\mZ\in \R^{N\times C_f}$, where the $C_f$ is the feature dimension.
The process is as follows:  
\begin{equation}
  \mZ=\mathrm{ReLU}(\mA \mX \mW), 
\end{equation}
where $\mW \in \R^{C_n \times C_f}$ is the parameter of GCN.  
The global adjacency matrix $\mA$ is continuously updated using multiple observations. 

\begin{figure}
  \centering
  \includegraphics[width=1\linewidth]{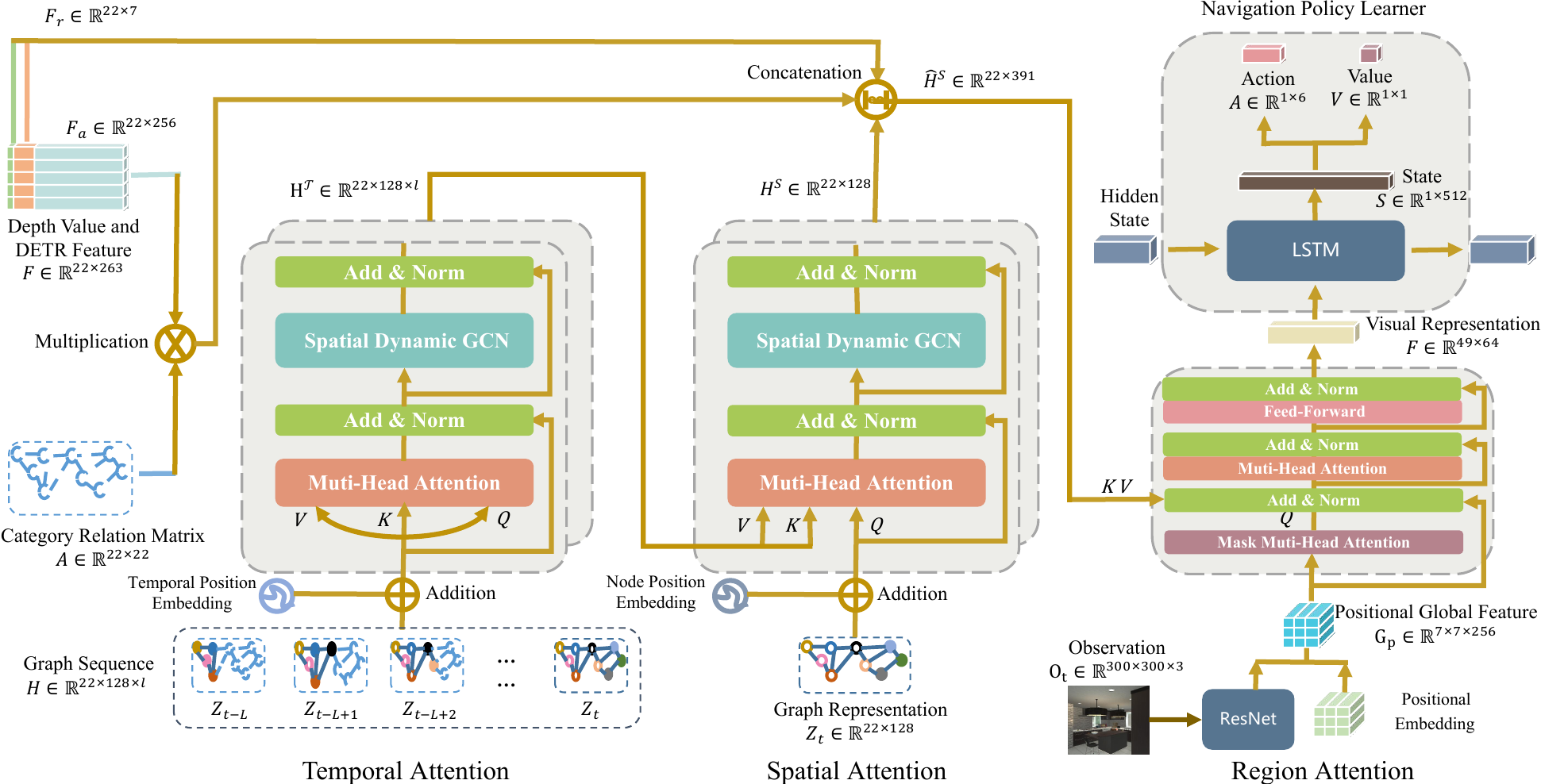}
  \caption{\textbf{Overview of TSR attention architecture.}
  Based on the category relation graph sequence, we propose a temporal attention module to model the temporal structure, which implicitly encodes the historical moving or trajectory information. 
  Then, we introduce a spatial attention module to uncover the spatial context of the current observation based on the category relation graph and past observation representation. 
  Third, we utilize a region attention module that shifts the attention to the target-relevant region and obtains the final visual representation. 
  Finally, a standard A3C architecture is implemented to learn the navigation policy.
  }
  \label{HTORG_model}
\end{figure}

Because not all nodes are observable in the multiple observations or utilized to learn, we only use the observable category nodes in each observation. 
For the specific $j$-th categories node feature $x_j$ in the category relation graph, if the category object does not exist in the current observation, we set the node feature $x_j$ to \textbf{0}. 
Based on $\mA$ and the observation, we can build a graph to describe the relations of the object categories in the current scene.   
Specifically, for the $i$-th object category node of the graph, our process of learning the node representation $\vz_i$ of $\mZ$ is formulated as follows:
\begin{equation}
\vx_i=\left\{
             \begin{array}{lr}
             \vx_j,  & \vx_j\in O
             \\0,&  \vx_j \notin O 
             \end{array},
\right.
\vz_i=[\sum_j^{N} a_{ij}\vx_j]\mW,
\end{equation}
where $a_{ij}$ denotes the adjacency value between node $i$ and node $j$ in adjacency matrix $\mA$.
Then, we construct a graph representation sequence 
$\tH = (\mZ_{t-l},\mZ_{t-l+1},\mZ_{t-l+2},\cdots,\mZ_{t}) \in \mathbb{R}^{N \times C_f \times L}$,
where $L$ denotes the length of the sequence. 
We feed $\tH$ into the Temporal-Spatial-Region attention architecture (TSR) and learn the navigation policy, as shown in Figure \ref{HTORG_model}. 

\subsection{Temporal Attention}
We design a temporal attention module to capture the structure of object relations among multiple graph representations. 
The important layer in this module is Multi-Head Attention (MHA)~\cite{vaswani2017attention} and Spatial Dynamic GCN (SDGCN)~\cite{guo2021learning}. 
For the $n$-th category, we feed its representation 
$\mH_n \in \R^{C_f \times L}$ into the MHA and perform self-attention. 
We add the temporal-position embedding to the sequence to emphasize the positional information for different temporal steps.
The process of self-attention is as follows: 
\begin{equation}
    \mH^{\gT}_n=\mathrm{Softmax}(\frac{\mH_n\mH_n^\top}{\sqrt{C_f} })\mH_n. 
\end{equation}
Then, we employ SDGCN to embed  $\{\mH^{\gT}_1, \cdots, \mH^{\gT}_N\}$ and update node representations; in this way, representations can capture the category relation across node dimensions.
After the temporal attention module, the representation $\mH^{\gT}$ can focus on the history observations.

\subsection{Spatial Attention} 
The spatial attention module emphasizes the object layout under the current observation $\tO_{t}$.
Similar to temporal attention, the spatial attention layer also consists of an MHA and an SDGCN.
Specifically, this MHA employs $\{\mH^{\gT}_1, \cdots, \mH^{\gT}_N\}$ as the key and value,
and the graph representation  $\mZ_t$ is the query.
To emphasize the different object nodes, we utilize the node-position embedding to add the graph nodes before the attention.
Each node $\mH^{\gS}_n$ of the graph representation in the multi-head attention update process is as follows:
\begin{equation}
\mH^{\gS}_n=\mathrm{Softmax}(\frac{{\mZ_{t,n}}(\mH^{\gT}_n)^\top}{\sqrt{C_f} })\mH^{\gT}_n.
\end{equation}
After this, SDGCN is also utilized to update the graph node representation, then the category relation graph representation of all nodes is $\mH^{\gS}$.

To make the agent pay more attention to the appearance features, we apply the category relation matrix $\mA$ as the attention map to the DETR appearance feature  $\mF_{a}$.
Specifically, we perform a matrix multiplication of $\mF_{a}$ and the matrix $\mA$, then feed to a nonlinear transformation layer $ReLU()$.
Following the ResNet18 \cite{he2016deep}, we concatenate the graph representation $\mH^{\gS}$, the emphasized appearance feature, and the other detection features to prevent the information from missing.
The concatenation process is as follows:
\begin{equation}
\hat{\mH}^{S}=\left|\mathrm{ReLU}(\mF_{a}\mA), \mF_{r},\mH^{\gS} \right|,
\end{equation}
where $\left|\cdot, \cdot \right|$ denotes concatenate operation, 
$\mF_{r} \in \mathbb{R}^{ N \times C_r}$ includes the bounding box $[x_1,y_1;x_2,y_2]$,
the detection confidence value $c$, the depth value $d$, and the semantic label $s$.

\subsection{Region Attention} 
The region attention module associates the graph representation with the visual observation regions.
It consists of a masked multi-head attention, a multi-head attention, and a feed-forward layer.
Following~\cite{DBLP:conf/iclr/DuY021}, we adopt ResNet18 and positional embedding to obtain the positional global feature $\tG_{p} \in \mathbb{R}^{H\times W \times C_g}$ of the observations $\tO_t$.
For attention to the graph representation, we reshape $\tG_{p}$ into a matrix-based representation $\mG_{p} \in \mathbb{R}^{HW \times C_g}$ and utilize the Multi-Layer Perceptron (MLP) to update the shape of $\hat{\mH}^{\gS}$ to $\mathbb{R}^{N \times C_g}$.
We utilize the masked multi-head attention to embed the positional global feature $\tG_{p}$, then utilize it as the query for the later multi-head attention.
We feed the category graph representation $\hat{\mH}^{\gS}$ into the attention as key and value.
The multi-head attention function is expressed as:
\begin{equation}
\mF = \mathrm{Softmax}(\frac{\mG_{p}(\hat{\mH}^{\gS})^\top}{\sqrt{C_g}})\hat{\mH}^{\gS}. 
\end{equation}
The feed-forward layer is utilized to further process the visual representation.
After all the processing of the region attention module, we get $\mF$ as the final informative visual representation.

\subsection{Policy Learning and Pre-training}
Based on the visual representation $\mF$, we implement a standard Asynchronous Advantage Actor-Critic (A3C) architecture \cite{DBLP:conf/iclr/BabaeizadehFTCK17} as the policy learner.
Specifically, we employ a standard Long Short Term Memory (LSTM) \cite{hochreiter1997long} network to fuse the previous hidden states and the input representation $\mF$.
In each worker of A3C, the policy function is called Actor to output the action distribution, and the value function is called Critic to generate the action value.
Unlike supervised learning, this part does not have a dataset but a simulated interactive environment.
The agent selects the action with the highest probability value and executes it in the simulation environment.
The environment returns a reward value and next state to the agent, and then the agent utilizes the experience data to learn the navigation policy.

Before formal training, we introduce a pre-training stage to accelerate the convergence of our representation extractor network.
In this part, without the navigation policy learner, we directly utilize an MLP to predict action distributions based on the observation sequence.
We utilize Dijkstra's algorithm as the expert policy to interact with the environment, generating the pre-train trajectory dataset.
We feed the observations trajectory $\gT = \{\tO_t, \cdots, \tO_{t+L}\}$ to the representation extractor network and utilize the optimal action as the trajectory label.
We implement a cross-entropy loss $L_{pre}=CE(f(\gT),\hat{\va})$ to train our representation extractor network, where $\hat{\va}$ represents the optimal action generated by Dijkstra's algorithm.

\section{Experiments}
\subsection{Experiment Setup}

We train and evaluate our CRG-TSR using the AI2-THOR \cite{kolve2017ai2} simulator, a large-scale interactive simulation environment for Embodied AI.
It provides near photo-realistic observation in 3D indoor scenes of four rooms, \textit{i.e.,} bedroom, kitchen, living room, and bathroom. 
Each scene contains 30 rooms with unique spatial layouts, object types, and appearances.
The initialized state is randomly chosen from over 2000 states.
Following the setting in \cite{DBLP:conf/iclr/DuY021}, we choose 20 scenes in each room type for training, 5 for validation, and 5 for testing.
The validation and testing rooms are unknown to the agent, and compared to the training rooms, they contain the same object categories but different object layouts and room styles.
We also consider 22 types of objects and ensure at least 4 potential targets for each scene. 
The step size of MoveAhead is 0.25 meters, and the angles of turning left/right and looking up/down are 45$^{\circ}$ and 30$^{\circ}$, respectively. 
During training, the maximal episode length is $50$; during validation and testing, the maximal length is limited to $100$ in living rooms and $50$ in other room types.

\subsection{Evaluation Metrics}
We adopt the Success Rate (SR) and Success Weighted by Path Length (SPL) \cite{DBLP:journals/corr/abs-1807-06757} as evaluation metrics to evaluate our method.
SR is utilized to measure the effectiveness of navigation.
It is formulated as $\frac{1}{N_{e}}\sum_{n=0}^{N_{e}}S_{n}$, where $N_{e}$ represents the total number of episodes, and $S_{n}$ is a binary indicator that defines the success of the $n_{th}$ episode.
Specifically, 1 represents a successful episode, while 0 denotes that the episode fails.
SPL represents the efficiency of navigation trajectories, considering both success rate and path length.
It is defined as $\frac{1}{N_{e}}\sum_{n=0}^{N_{e}}S_{n}\frac{L_{opt}}{max(L_{n},L_{opt}))}$, where $L_{n}$ and $L_{opt}$ represent the path length and the optimal path length of the current episode, respectively.
More experimental and visualization results are in the supplementary material.

\subsection{Comparison Methods}
We compare several classic algorithms in the AI2-THOR environment as follows:
{\bf Random} refers to the agent choosing actions based on the average action probability. 
{\bf WE}  \cite{pennington2014glove} adopts Glove embedding to represent the target category.
{\bf SP} \cite{DBLP:conf/eacl/GraveMJB17} learns the category relationship among the objects from the external knowledge data FAstText \cite{DBLP:conf/eacl/GraveMJB17} and employs scene prior knowledge to navigate.
{\bf SAVN} \cite{wortsman2019learning} is a meta-reinforcement learning method that allows agents to adapt quickly to unseen environments. 
{\bf ORG} \cite{du2020learning} employs the Faster RCNN detection feature to establish an object relation graph from each frame.
ORG+TPN is the ORG adding a memory-augmented tentative policy network (TPN) module to escape deadlocks. 
{\bf OMT \cite{DBLP:conf/icra/FukushimaOKSY22}} enables the store of long-term scenes and object semantics and attends to salient objects in the sequence of previously observed scenes and objects. 
{\bf HOZ} \cite{zhang2021hierarchical} utilizes the Faster RCNN feature to construct an online-learning hierarchical object-to-zone graph, and the agent can constantly plan an optimal path from zone to zone.
{\bf VTNet} \cite{DBLP:conf/iclr/DuY021} implies a Transformer to fuse two spatial-aware descriptors and achieve the visual representation.
{\bf SSCNav \cite{DBLP:conf/icra/LiangCS21}} employs a confidence-aware module to model the scene and guide the agent's navigation planning.
{\bf ACRG}  \cite{hu2023agent} 
proposes an agent-centric relation graph and analyzes the different roles of object relationships in different directions for navigation.
{\bf SHRL}  \cite{wang2023skill} proposes a skill-based hierarchical method containing a high-level policy and three low-level skills. 
It utilizes high-level policy to select low-level skills and achieve navigation tasks.
{\bf DOA \cite{DBLP:conf/mm/DangSWHLC22}} is a directed object attention graph guiding the agent learning the attention relationships between objects without the object attention bias.
{\bf PONI \cite{ramakrishnan2022poni}} predicts two complementary latent functions conditioned on the semantic graph and utilizes them to decide where to look for objects.
{\bf Baseline } directly implements the concatenation of the DETR detection and global features to A3C for policy learning. 
It is a vanilla version of our method.

\subsection{Training Details}
We follow the work \cite{DBLP:conf/ijcai/ChenCZ19} utilizing the NYU-Depth v2 dataset \cite{silberman2012indoor} to pre-train the depth estimation model. 
We utilize the ResNet18 \cite{he2016deep} pre-trained on ImageNet \cite{deng2009imagenet} as our backbone to extract the features of each observation $\tens{O}$.
In the pre-training stage, we adopt a learning rate decay strategy for supervised learning, starting from the initial learning rate $L_{int} = 10^{-5}$.
The Dijkstra's expert policy is utilized to generate the pre-training observation sequence dataset.
We utilize the DETR \cite{carion2020end} as the object detector rather than Faster RCNN \cite{faster2015towards} or the one-stage method RTMDet \cite{lyu2022rtmdet}.  
The feature output by the DETR detector is embedded with position-encoded global context information, which is more suitable for feature fusion operations than Faster RCNN.
For RTMDet, it cannot capture specific object appearance features and is also unsuitable.
Finally, we train the policy with 16 asynchronous workers in 2M navigation episodes and utilize the Adam optimizer \cite{DBLP:journals/corr/KingmaB14} to update network parameters with a learning rate of $10^{-4}$.
The number of layers in each attention module is $2$, the number of heads in the multi-head attention is $4$, and the representation sequence length is $4$.
The average time for a single agent to process each image is 0.07 seconds, and it can process 13 images in one second.
We conduct model training and testing under the Unbutu 20.04.1 system. The CPU version is Hygon C86 7151 16-core Processor, and the CPU clock frequency is 1800MHZ.
The graphics card we use is NVIDIA RTX A4000, the driver version is 470.82.00, and the Cuda version is 11.4.
During the training process, we use 16 agents on 4 graphics cards to train 2 million episodes; each agent process occupies 3043MiB of memory, and the total training time is 45 hours.

\begin{table*}
\centering
\caption{\textbf{Comparison with other navigation models.} We report the  Success Rate (SR), Success Weighted by Path Length (SPL), and their variances in parentheses by repeating experiments five times. 
}
\resizebox{0.8\textwidth}{!}{
\begin{tabular}{p{3cm}|c|cc|cc}
\hline
\multirow{2}{*}{\textbf{Method}} & \multirow{2}{*}{\textbf{Episodes}} & \multicolumn{2}{c}{\textbf{ALL}} \vline  & \multicolumn{2}{c}{$L_{opt}\geqslant 5$}\\[1pt] \cline{3-6}
                     &  & \makecell[c]{\textbf{SR}} & \makecell[c]{\textbf{SPL}}   &   \makecell[c]{\textbf{SR}}  & \makecell[c]{\textbf{SPL}}\\[1pt]\hline
\hline
    Random    &-  &8.0(1.3)    &0.036(0.006)  &0.3(0.1)   &0.001(0.001)  \\
    WE \cite{pennington2014glove}         &-  &33.0(3.5) &0.147(0.018)  &21.4(3.0)  &0.117(0.019)  \\
    SP \cite{DBLP:conf/iclr/YangWFGM19}         &- &35.1(1.3) &0.155(0.011)  &22.2(2.7)  &0.114(0.016)  \\
    SAVN  \cite{wortsman2019learning}       &- &40.8(1.2)  &0.161(0.005)  &28.7(1.5)  &0.139(0.005) \\
    ORG \cite{du2020learning}     &6M &65.3(0.7)  &0.375(0.008)  &54.8(1.0)  &0.361(0.009) \\
    ORG+TPN \cite{du2020learning}    &6M &69.3(1.2)  &0.394(0.010) &60.7(1.3) &0.386(0.011)  \\
    OMT \cite{DBLP:conf/icra/FukushimaOKSY22} &- &70.5(-) &0.304(-) &59.8(-)&0.293(-) \\
    HOZ  \cite{zhang2021hierarchical}       &6M      &70.6(-)   &0.400(-)  &62.7(-)  &0.392(-)  \\ 
    VTNet \cite{DBLP:conf/iclr/DuY021}     &6M  &72.2(1.0)  &0.449(0.007)  &63.4(1.1)  &0.440(0.009)  \\
    SSCNav \cite{DBLP:conf/icra/LiangCS21} &- &76.5(-) &0.244(-) &68.1(-) &0.299(-) \\
    ACRG \cite{hu2023agent}   &2M &77.6(1.1)  &0.439(0.012)  &71.0(0.5)  &0.423(0.007)  \\
    SHRL \cite{wang2023skill}   &6M &78.3(0.8)  &0.418(0.005)  &70.6(0.7)  &0.415(0.005)  \\
    DOA \cite{DBLP:conf/mm/DangSWHLC22}   &- &78.8(-) &0.443(-) &72.4(-) &\textbf{0.455(-)} \\
    PONI \cite{ramakrishnan2022poni}  &- &79.6(-)& 0.356(-)& 73.3(-)& 0.364(-) \\
\hline 
Baseline  &6M    &66.2(0.6)  &0.352(0.007)  &54.0(0.5)  &0.336(0.006) \\ 
\textbf{CRG-TSR (Ours)}   &2M  &\textbf{80.0(0.8)}  &\textbf{0.456(0.005)}  &\textbf{73.6(1.1)}  &0.451(0.007)  \\\hline
\end{tabular}}
\label{mainly}
\end{table*}

\subsection{Comparison with Competing Methods}

\subsubsection{Improvement over Baseline}
Our method surpasses the baseline by a large margin on both SR ($+20.8\%$) and SPL ($+29.5\%$), as shown in Table \ref{mainly}. 
Baseline only directly utilizes the DETR detection and global features for navigation policy learning without modeling the relationship between categories and perceiving the historical memory.
This comparison result illustrates the effectiveness of our CRG-TSR method.

\subsubsection{Comparison with Related Arts}
As shown in Table \ref{mainly}, our CRG-TSR significantly outperforms the state-of-the-art methods,
improving the SR from 79.6\% (PONI) to 80.0\%, SPL from 0.449  (VTNet) to 0.456, train episodes from 6M to 2M, and our algorithm achieves the joint optimality of all metrics simultaneously.
These experimental results denote that our method does capture a better visual representation and increase the navigation's effectiveness and efficiency.

Our method outperforms the classical algorithms by a large margin.
SP utilizes external knowledge data to encode category relationships and cannot adapt to specific environments.
SAVN directly connects features from various modalities, but the gap between different modalities may not facilitate learning.
ORG, ACRG, and VTNet establish the correlation between objects and objects or objects and the observation regions differently.
However, they only consider the observation image of a single observation and ignore the information of multiple observations in historical trajectories.
OMT considers the impact of long-term memory on navigation and introduces a memory storage module to prevent memory forgetting but does not conduct an in-depth analysis of the stored object memory to serve current decisions.
HOZ depends on prior knowledge to construct a global object-to-zone graph, but the modeling of regions is rougher than the modeling of object information and cannot provide clear guidance of object information.
SSCNav takes a lot of extra steps to model the scene's semantical map. 
Although the navigation effect is improved somewhat, it greatly reduces efficiency.

Our method is also ahead of the current state-of-the-art algorithms.
SHRL is a skill-based hierarchical reinforcement learning method.
It proposes a complex hierarchical policy learning method but does not consider the agent's state awareness of the environment. 
Due to the lack of global perception of the scene, it is still easy to fall into the local optimal solution.
DOA considers the problem of invisible objects in single-frame observations and proposes to use directed graphs to solve the confidence bias problem. 
Unlike it, our method leverages trajectory information to capture global object layout details, enabling the agent to perceive invisible objects and comprehensively understand scene information.
PONI is the state-of-the-art model navigation method, decoupling navigation actions and perceptions.
However, the SPL of PONI is very poor because it lacks the correlation information between specific navigation actions and scene perception.
Therefore, although the future object position can be predicted in the environment, there may be many attempted actions for PONI.

Our method adaptively learns the relationship information between object category layouts in the training environment and can generalize to unknown layout scenarios faster and better.
We also propose three attention modules to capture the long-term spatio-temporal dependencies of the graph representation sequence. 
Our model can establish the relationship between the graph representation sequence and the global feature of the scene observation.
Our representation processing is more detailed and more in line with human navigation, having higher efficiency and effectiveness.

\begin{figure}
  \centering
  \includegraphics[width=0.8\linewidth]{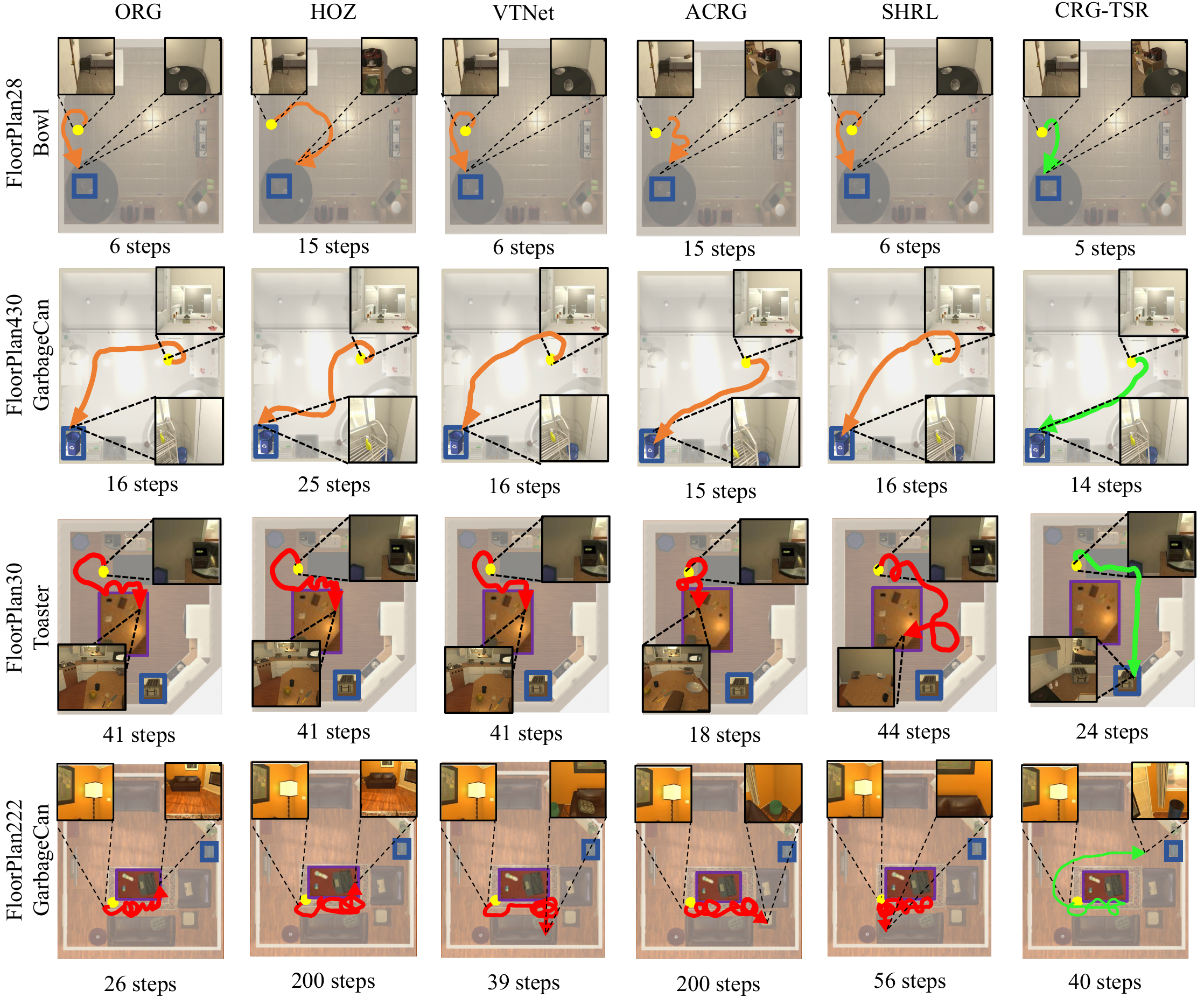}
  \caption{\textbf{Visualization results of different models in the test environment. }
  The target object is highlighted with blue bounding boxes,  the obstacle is highlighted with purple bounding boxes, and the starting point is marked with a yellow dot.
  The green line represents the best case, the orange line represents the sub-optimal case, and the red represents the failed case. 
  We show four room categories and target objects, then label the room number and target category on the left side of each row.
  }
  \label{case_study}
\end{figure}

\subsubsection{Qualitative Results}
To further vividly demonstrate the excellence of the navigation performance of our algorithm, we selected five open-source models in the AI2-THOR environment for navigation trajectories case analysis, \textit{i.e.}, ORG, HOZ, VTNet, ACRG, and SHRL.
As shown in Figure \ref{case_study}, we chose four navigation rooms and two class tasks, \textit{i.e.}, ``Easy'': anyone can do, and ``difficult'': only ours can do.


As shown in the first two rows of Figure \ref{case_study}, five approaches complete the task, but our method implements a shorter path.
Analyzing the trajectory direction, our model, HOZ, and ACRG tend to move to the area with many objects.
However, ORG, VTNet, and SHRL tend to explore from the other direction.
In a specific comparison of HOZ and ACRG, our method takes fewer steps.
The results show that our category relation graph is more efficient than the online-learning hierarchical object-to-zone graph of HOZ and the different direction relationships graph of ACRG.

In the last two rows of Figure \ref{case_study}, the task becomes more difficult due to the obstacle (table) in the path.
Except for our CRG-TSR, the remaining methods fail. 
ORG, VTNet, and ACRG can find the target object and move toward it but fail because of the obstacle.
These methods only utilize single-frame observation as input, so they easily forget historical information and fall into the obstacle deadlock state.
HOZ tries to find a path to escape obstacles but fails due to the limit of the number of steps.
Although SHRL can escape from obstacles by choosing a random exploration policy hierarchically, it is easy to fall into a new deadlock state.
However, our method can successfully escape deadlocks and reach the target object.
We learn a category relation graph as knowledge so the agent can better infer the target object's position and perceive the scene's layout to escape from the obstacle.
At the same time, the trajectory after leaving the obstacle shows that our agent still remembers the position of the target object and can move toward the target quickly.
It illustrates the importance of the historical memory built by the TSR attention mechanism.

\begin{table}
\begin{center}
\caption{\textbf{Impacts of different components on navigation performances.}
The left indicates the usage of each component in the model, and the right is the report result including the SR, SPL, and their variances. 
\checkmark is displayed the component is used.
}
\resizebox{0.8\textwidth}{!}
{
\begin{tabular}{C{0.8cm}C{1.6cm}C{1.2cm}C{1.4cm}|cc|cc}
\hline
\multicolumn{4}{c}{\textbf{Method}}\vline  & \multicolumn{2}{c}{\textbf{ALL}} \vline& \multicolumn{2}{c}{$L_{opt}\geqslant 5$}\\[1pt] \hline
\textbf{CRG}  & \textbf{Temporal}&\textbf{Spatial} & \textbf{Region} &    
                        \makecell[c]{\textbf{SR}} & \makecell[c]{\textbf{SPL}}     & \makecell[c]{\textbf{SR}} & \makecell[c]{\textbf{SPL}}\\[1pt]\hline
\hline
 & & & &66.2(0.216)  &0.352(0.007) &54.0(0.464) &0.336(0.006)\\[1pt]
\checkmark &  &  &   &75.2(0.386)   &0.408(0.006)     &66.4(0.787)  &0.407(0.006) \\[1pt]
&\checkmark &\checkmark &\checkmark &76.4(0.793)   &0.452(0.006)  &68.5(1.067)  &0.437(0.007)  \\[1pt]
 \checkmark & \checkmark  &\checkmark   &      &76.5(0.249)   &0.419(0.004) &68.2(0.647)  &0.402(0.005)\\[1pt]
\checkmark  &\checkmark   &  &   \checkmark    &77.0(0.432)   &0.426(0.005) &68.8(0.612)  &0.415(0.008)\\[1pt] 
\checkmark  &  &\checkmark   &\checkmark       &77.1(0.374)   &0.444(0.005) &69.5(0.125)  &0.437(0.005) \\[1pt]\hline
\hline
\checkmark  &\checkmark  &\checkmark   &\checkmark       &\textbf{80.0(0.825)}   &\textbf{0.456(0.005}) &\textbf{73.6(1.131)}  &\textbf{0.451(0.007)} \\[1pt]\hline
\end{tabular}}
\label{alnby}
\end{center}
\end{table}

\subsection{Ablation Study}
In Table \ref{alnby}, we provide experimental results to validate the contribution of CRG and TSR by adding individual ones to the baseline framework.

When CRG is adopted, overall performance is significantly improved.
SR increases from 66.2\% to 75.2\% and SPL from 0.352 to 0.408.
Similarly, using TSR also improves performance, \textit{i.e.,} SR increases from 66.2\% to 76.4\% and SPL from 0.352 to 0.452.
The detailed results indicate that each technique of our method is effective, and the attention structure has a greater impact on navigation efficiency.
However, compared with our full model, only learning the category relation or modeling the historical memory by attention does not work well.
Our CRG-TSR method degrades if only one aspect is considered.
In a word, both parts of our work are indispensable and complementary. 

The role of the different attentions in the TSR mechanism is also shown in Table \ref{alnby}.
After removing the individual components, the results display that every attention module is indispensable.
The temporal and spatial attention modules are utilized for encoding the graph representation sequence and emphasizing the object layout under the observation.
Without these modules, the agent cannot sufficiently perceive the positional relationships of objects, and the navigation SR will degenerate from 80\% to 77\%.
The region attention module establishes an association between graph representation and image regions.
Only utilizing graph relationships and abandoning the perception of the image regions leads to larger performance degradation from 80\% to 76.5\%.
The result shows that the perception of image regions is also important.

\subsection{Variant Study}
We fully study the impact of different modules in our method on navigation effects, such as sequence length, attention module architecture, pre-training stages, object detection methods, and graph construction methods.

\subsubsection{Impact of Sequence Length}
The length $L$ of the graph representation sequence influences the performance.
Table \ref{hyper} shows that SR initially increases and then decreases as the value of $L$ increases, reaching its peak performance when $L$ is set to $4$.
When $L$ is small, the attention modules cannot better consider the long-term spatio-temporal dependencies of the sequence.
Conversely, if the number is large, more trajectory memory and noise may be introduced when capturing the object relationships.

\begin{table*}
\centering
\caption{\textbf{Comparison of different sequence lengths $L$.} 
We verify the effect of different sequence lengths and report the results.
The best performance is achieved by setting the sequence length $L=4$.}
\resizebox{0.8\textwidth}{!}{
\begin{tabular}{c|cc|cc}
\hline
\multirow{2}{*}{\textbf{Method}}  & \multicolumn{2}{c}{\textbf{ALL}} \vline& \multicolumn{2}{c}{$L_{opt}\geqslant 5$}\\[1pt] \cline{2-5}
                        & \makecell[c]{\textbf{SR}} & \makecell[c]{\textbf{SPL}}     & \makecell[c]{\textbf{SR}} & \makecell[c]{\textbf{SPL}}\\[1pt]\hline
\hline
$L=1$ &77.2(0.782)   &0.449(0.006) &69.8(0.660)  &0.440(0.006)\\[1pt]
$L=2$ &78.6(0.936)   &0.444(0.004) &70.9(1.158)  &0.445(0.008)\\[1pt] 
$L=4$ &\textbf{80.0(0.825)}  &\textbf{0.456(0.005)} &\textbf{73.6(1.131)}  &\textbf{0.451(0.007)}  \\
$L=6$ &77.9(0.582) &0.447(0.007) &69.9(0.660)  &0.443(0.011) \\[1pt]
$L=8$ &78.0(0.656) &0.451(0.002) &70.0(0.368)  &0.445(0.002)  \\[1pt]\hline
\end{tabular}}
\label{hyper}
\end{table*}

\begin{table*}
\begin{center}
\setlength\tabcolsep{2.5pt}
\caption{\textbf{Comparison of different attention architectures.} We report the pre-training accuracy on the validation dataset and navigation results on the testing scene.
}
\label{number}
\resizebox{0.8\textwidth}{!}{
\begin{tabular}{c|c|c|cc|cc}
\hline
\multirow{2}{*}{\shortstack{Head\\Number}} & 
\multirow{2}{*}{\shortstack{Attention\\Layer}} & 
\multirow{2}{*}{Accuracy}&
\multicolumn{2}{c}{ALL} \vline  & \multicolumn{2}{c}{$L_{opt}\geqslant 5$}\\[1pt] \cline{4-7}
                     &  &  & \makecell[c]{SR} & \makecell[c]{SPL}   &   \makecell[c]{SR}  & \makecell[c]{SPL}\\[1pt]\hline
\hline
\multirow{3}{*}{\textbf{4}}  
& 1 &0.944 &78.5(0.556) &0.440(0.009) &71.0(0.818) &0.437(0.008)    \\[1pt]
&\textbf{2} &\textbf{0.971} &\textbf{80.0(0.825)} &\textbf{0.456(0.005) }&\textbf{73.6(1.131)} &\textbf{0.451(0.007) }   \\[1pt]
&4 &0.956 &79.5(0.834)  &0.463(0.004) &72.7(0.779) &0.451(0.003) \\[1pt]\hline
\multirow{3}{*}{8}  
&1 &0.937 &78.5(0.245)  &0.458(0.003) &71.0(0.403) &0.450(0.004)    \\[1pt]
&2 &0.969 &79.2(0.460)  &0.447(0.007) &71.6(0.563) &0.439(0.009)    \\[1pt]
&4 &0.962 &77.8(0.170)  &0.445(0.007) &69.6(0.283) &0.437(0.004)   \\[1pt]
\hline 
\end{tabular}}
\end{center}
\end{table*}

\subsubsection{Impact of Attention Architectures}
We compare the effect of different attention architectures on model performance in Table \ref{number}.
We vary the number of heads in multi-head self-attention and the number of layers in all attention modules.
Specifically, the three attention modules in TSR modules have the same number of layers and heads.
We have yet to do an exhaustive search so that better combinations may exist.
We observe that the model fails to converge to an optimal policy when the attention modules become too deep. 
In contrast, the attention module with single layers does not have enough network parameters to generate representative visual features.
The highest SR is achieved when the attention module contains 4 heads and 2 layers.

\subsubsection{Impact of Pre-training stages} 
\begin{figure}
  \centering
  \includegraphics[width=0.8\linewidth]{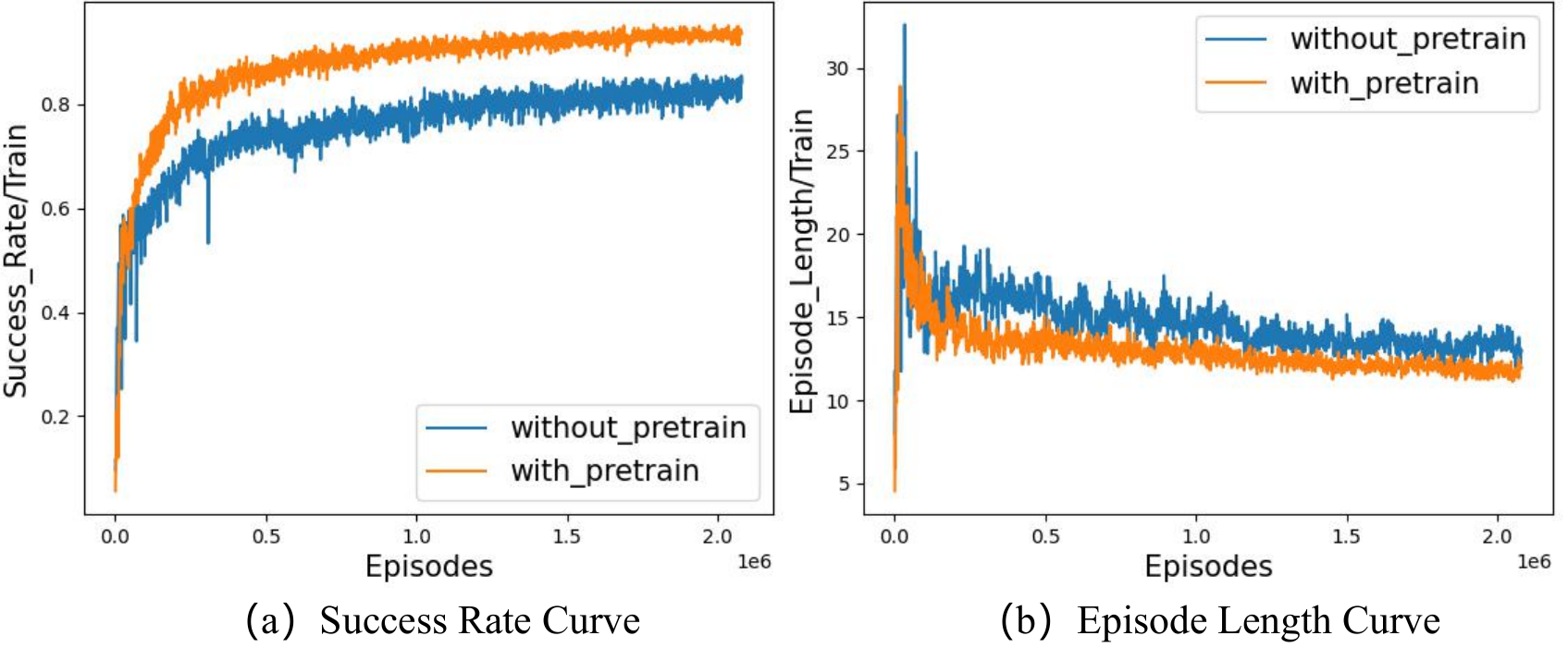}
  \caption{\textbf{Visualization of the pre-training phase importance.}  We compare our model with and without the pre-training scheme. Orange and blue curves represent with and without pre-training, respectively.
  }
  \label{pretrain_cur}
\end{figure}

To illustrate the implications of pre-training, we compared the curves of our model with and without the pre-training phase during training in Figure \ref{pretrain_cur}.
Specifically, the success rate curve is compared in (a), and the episode length curve is compared in (b).
The results show qualitatively that the pre-training stage speeds up parameter convergence.
For the success rate curve, our model with pre-training achieves higher sample efficiency and performance than the model without pre-training in 1 million training episodes.
Although the model without pre-training improves performance as the training episodes increase, the speed is slower than the model with the pre-training phase.
For the episode length curve, the model with pre-training improves navigation efficiency and spends nearly 13 steps per episode in training. 
In contrast, the model without pre-training slows convergence and has more steps to achieve navigation.

\subsubsection{Impact of Object Detection Method}
To achieve a fairer comparison with previous algorithms, we conducted variant experiments using the Faster RCNN algorithm, and the results are shown in Table \ref{Variants}.
While a superior object detection algorithm can boost performance, our algorithm's performance still surpasses previous algorithms even when using the same object detector. 
For example, ORG and HOZ also utilized the Faster RCNN algorithm, yet our navigation success rate reached 76, while their algorithms only achieved 65.3 and 70.6, respectively. 
Additionally, for SR of these models employing the DETR, such as  VTNet (72.2), ACRG (77.6), and SHRL (78.3), our model utilizes DETR and achieves optimal results on navigation performance.

\subsubsection{Impact of Graph Construction Method}
To validate the effectiveness of CRG, we compare different variants for graph construction methods and experiment results in Table \ref{Variants}.
``Graph (distance)'' means using distance information between objects to represent the layout relationship \cite{li2018diffusion}.
``Graph (Recurrent)'' represents utilizing an inductive adaptive graph generation module to update the adjacency matrix \cite{Hu_2023}.
``Graph (Attention)'' represents utilizing the attention matrix to dynamically calculate the correlation between nodes \cite{9346058}.
The experimental results demonstrate the effectiveness of the CRG approach.
Our method utilizes a parameter dictionary to adaptively calculate the adjacency matrix and update the local adjacency relationship frame by frame, which is more suitable for learning object adjacency relationships in navigation tasks.

\begin{table*}
\centering
\caption{\textbf{Comparison of other different variants.}
We compare other variants of the object detection module and the graph construction method.
}
\resizebox{0.8\textwidth}{!}{
\begin{tabular}{p{4cm}|cc|cc}
\hline
\multirow{2}{*}{\textbf{Method}} & \multicolumn{2}{c}{\textbf{ALL}} \vline  & \multicolumn{2}{c}{$L_{opt}\geqslant 5$}\\[1pt] \cline{2-5}
                 &  \makecell[c]{\textbf{SR}} & \makecell[c]{\textbf{SPL}}   &   \makecell[c]{\textbf{SR}}  & \makecell[c]{\textbf{SPL}}\\[1pt]\hline
\hline
Detector (Faster RCNN)   &76.0(0.5)  &0.412(0.004) &70.7(0.5) &0.428(0.003) \\ 
Graph (Distance)  &78.3(0.1)   &0.447(0.008) &70.2(0.2)  &0.445(0.007)\\[1pt]
Graph (Attention)  &78.2(0.4)   &0.459(0.001) &70.4(0.5)  &0.448(0.001)\\[1pt] 
Graph (Recurrent)   &77.5(0.3)   &0.437(0.016) &70.0(0.2)  &0.435(0.011) \\[1pt]\hline
\hline
\textbf{CRG-TSR  }&\textbf{80.0(0.8)}  &\textbf{0.456(0.005) } &\textbf{73.6(1.1)} &\textbf{0.451(0.007) } \\\hline
\end{tabular}}
\label{Variants}
\end{table*}

\begin{table*}
\begin{center}
\caption{\textbf{Comparison of models on RoboTHOR environment.} We implement an offline version created based on an online environment and report the results. 
}
\label{Robothor}
\resizebox{0.5\textwidth}{!}{
\begin{tabular}{p{1.6cm} | cp{1cm} | cp{1cm}}
\toprule
Method  & \multicolumn{2}{c}{SR}\vline & \multicolumn{2}{c}{SPL}\\[1pt]\hline
\midrule
Baseline      &3.88  &(0.209) &0.0367 &(0.002) \\[1pt]
ORG \cite{du2020learning}      &9.97   &(0.676)  &0.0577  &(0.008)  \\[1pt]
HOZ \cite{zhang2021hierarchical}            &10.0   &(0.326) &0.0603  &(0.002)  \\[1pt] 
VTNet \cite{DBLP:conf/iclr/DuY021}            &10.4   &(0.346) &0.0655  &(0.001)  \\[1pt] 
ACRG \cite{hu2023agent}            &12.5   &(0.372) &\textbf{0.0687}  &\textbf{(0.003)}  \\[1pt] \hline 
\bottomrule 
\textbf{Ours}  &\textbf{16.2}   &\textbf{(0.591)}&0.0676  &(0.002)\\[1pt]\hline
\end{tabular}}
\end{center}
\end{table*}

\subsection{Results on RoboTHOR}
In this section, we discuss the performance of the navigation methods in a more complex environment RoboTHOR \cite{deitke2020robothor}.
The interaction between the agent and the environment in RoboTHOR is consistent with the AI2-THOR.
Following ACRG, we collect the observation data of the RoboTHOR environment and implement an offline version.
The input for the agent is only the observation data and no additional information.
Specifically, we selected 18 target categories and modified the maximal training episode length to $100$.
We choose 60 apartments for training, 5 for validation, and 10 for testing.
For a fair comparison, we implement six models on the RoboTHOR environment and guarantee all algorithms are compared under the same experimental settings. 
The results in Table \ref{Robothor} demonstrate their effectiveness.

The Baseline model is provided by the RoboTHOR ObjectNav 2021 Challenge.
In our setting, there is no prior knowledge of the scene, and all methods are trained for 2 million episodes.
Although the navigation task in Robothor is challenging due to several clapboards, our model outperforms other methods by a large margin, \textit{i.e.}, 5.8 on SR and 0.002 on SPL.
Specifically, ORG, VTNet, and ACRG perceive the target position directly through a single observation and easily forget the observation memory.
Since the regions in RoboTHOR are separated by clapboard and there without prior knowledge of the scene, HOZ has poor performance.

\begin{figure}
  \centering
\includegraphics[width=0.8\linewidth]{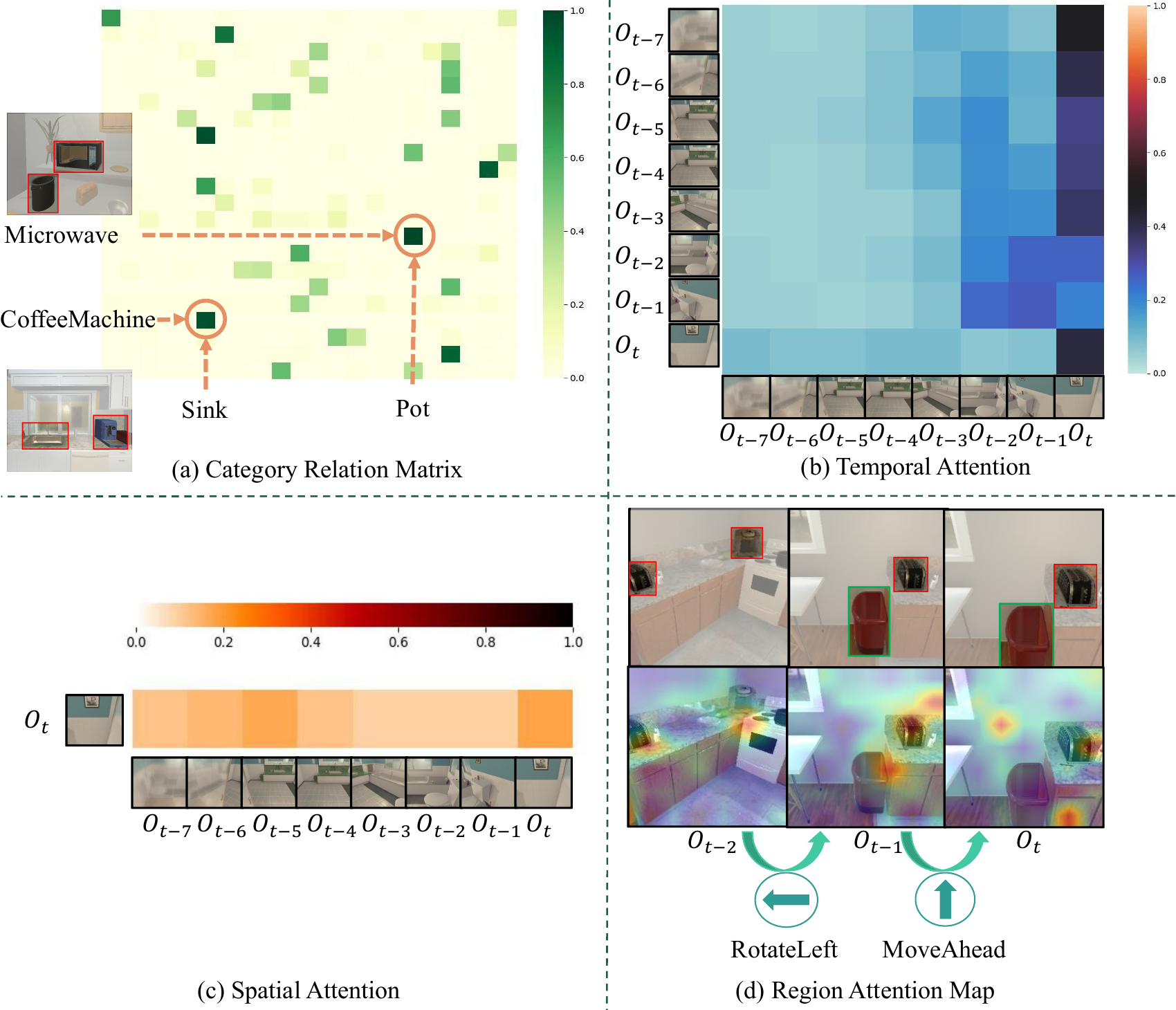}
  \caption{
  \textbf{Visualization of individual components in our method.} (a) visualizes the learned category adjacency matrix. 
  (b) and (c) show temporal and spatial attention on the trajectory observation, respectively. 
  (d) visualizes the actions and the corresponding region attention map of the observation sequence.
  }
  \label{expermentvisual}
\end{figure}

\subsection{Visualization Results}
\subsubsection{Category Relation Matrix}
During training, we learn a piece of knowledge about the relations of object categories in the world.
The relationship between object categories is represented using a category relation adjacency matrix, as depicted in Figure \ref{expermentvisual} (a). 
The rows and columns of the matrix correspond to the target object categories of interest. 
The column element with the highest weight in each row represents the category most closely related to that particular row. 
For instance, in the row corresponding to Microwave, the relationship with Pot is stronger than those with other categories. 
This indicates that the two objects frequently appear in the scene and are located close to each other. Similar cases can be observed for the Coffee-Machine and Sink.


\subsubsection{Temporal Attention}
To more clearly illustrate the mechanism of temporal attention, we set the length of the graph representation sequence $\tH$ equal to 8 and visualize the attention weights in Figure \ref{expermentvisual} (b).
The agent's latest observation is $\tO_t$, and its historical observations are $\tO_{t-1},\cdots, \tO_{t-7}$.
Although most historical representations focus on the last frame, the focus on other time steps is not 0.
This result demonstrates that although the latest observation is important for policy, historical information also plays an indispensable role.

\subsubsection{Spatial Attention}
We indicate the attention between the latest and historical observations for the spatial attention module shown in Figure \ref{expermentvisual} (c).
Same as Figure \ref{expermentvisual} (b), the agent's observation sequence is $\tO_{t},\cdots, \tO_{t-7}$.
Figure \ref{expermentvisual} (c) further reflects the importance of historical information.
Specifically, the latest observation $\tO_{t}$ not only pays more attention to itself but also pays attention to the early historical memory.
In this way, the agent can better perceive the latest observation and prevent forgetting historical information.

\subsubsection{Region Attention Map}
To demonstrate the effectiveness of the region attention module in associating navigation actions with the target-relevant regions, we visualize the attention map of the last encoder layer in the region attention module, as shown in Figure \ref{expermentvisual} (d).
The model focuses on a few target-relevant regions in the observation and then maps them to actions.
For the case in the first line, the object is visible, and our model directly detects the instances of interest and selects actions to approach the target image regions.
In contrast, the target objects are invisible in the case of the second and third lines. 
In these case, our model infers the approximate position of the target and attends to the regions in the image observation that corresponds to the moving direction.
Our CRG-TSR model endows agents with the ability to infer the location of target objects.
Guided by the observation attention map, the agent selects actions to approach the targets.

\subsection{Failure Case Study}
\begin{figure}
  \centering
  \includegraphics[width=0.7\linewidth]{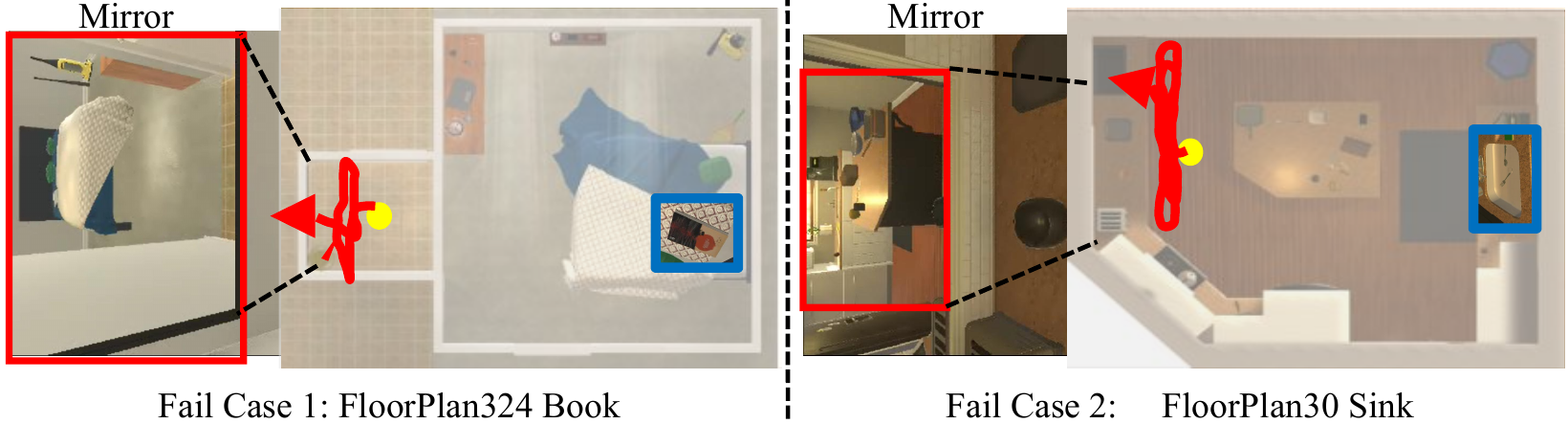}
  \caption{\textbf{Visualization results of failure cases.} 
Since our agent does not recognize the presence of specular reflections, it often fails to move toward the mirror.}
  \label{fail_case}
\end{figure}

As demonstrated in Figure \ref{fail_case}, our model fails to reach targets due to the specularity of the mirror.
In these two failure cases, the mirror reflects the room layout information. 
The agent lacks the perception of the mirror space, so the agent only wants to reach the area displayed in mirrors based on the observation and constantly moves to mirrors.
However, walking into the mirror space is impractical and leads to the agent's failure.
Specular reflections are frequently encountered in home environments, and the capability to distinguish specular reflections is a significant challenge for our future work in indoor visual navigation tasks.

\section{Conclusion}
This paper proposes an effective and efficient CRG-TSR method, utilizing the observation information in historical trajectory to solve the visual navigation task. 
We establish a Category Relation Graph (CRG) to learn certain layout knowledge about the relations of object categories.
Then, we propose a Temporal-Spatial-Region attention architecture (TSR) to further process the graph representation and obtain a final informative visual representation for policy learning.
Our method achieves superior navigation performance through systematic experimentation with quantitative and qualitative analysis compared to state-of-the-art methods.
The experimental results show that our visual representation is more informative and suitable for the navigation task.


\bibliographystyle{ACM-Reference-Format}
\bibliography{Reference}

\end{document}


\title{Supplementary Material}


\renewcommand{\shortauthors}{X.Hu et al.}

\received{7 July 2023}

\maketitle

This supplementary material provides more visualization of attention modules (Sec. \ref{sec:Visualization}) and more results on the different pre-training datasets (Sec. \ref{sec:Pre-training}).

\begin{figure}
  \centering
  \includegraphics[width=0.8\linewidth]{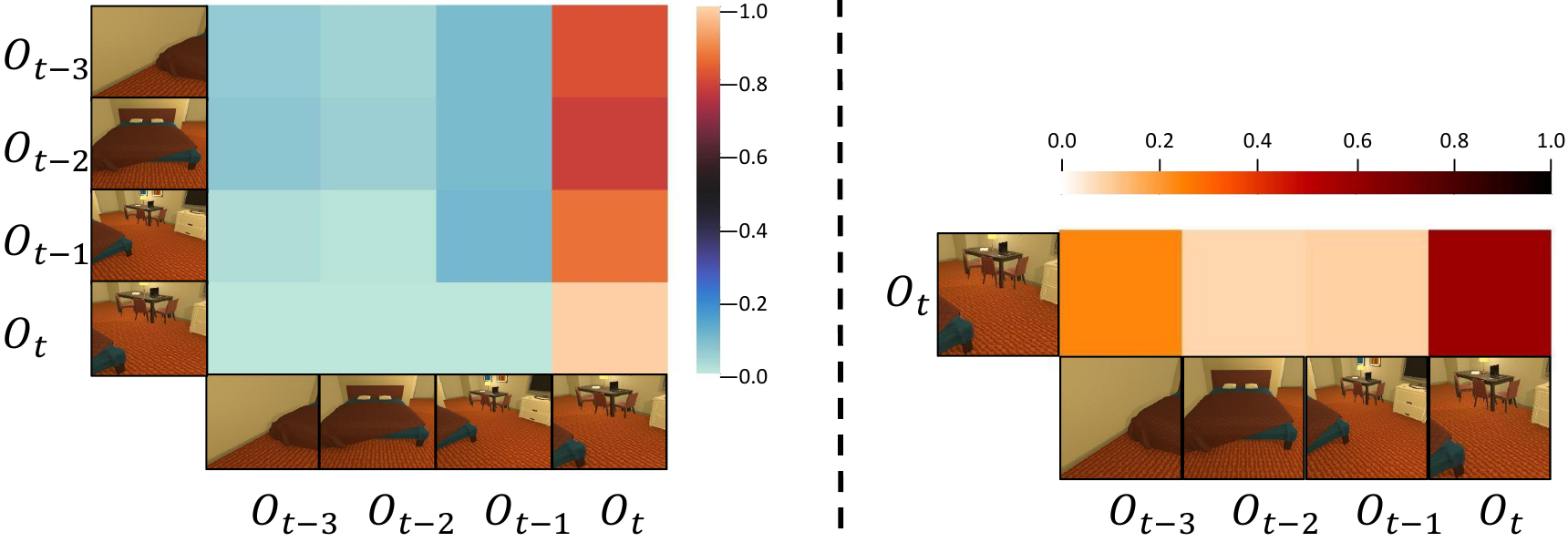}
  \caption{\textbf{Visualization of attention weights.} The left figure is the visualization of the temporal attention, and the right one is the spatial attention.
  moduleThis attention is between representation sequence elements. While most historical sequence representations focus on the last one, the focus on others is not zero.
  }
  \label{temproal_4}
\end{figure}

\section{More Visualization Results}
\label{sec:Visualization}

\subsection{More Temporal and Spatial Attentions}
We set the length of graph representation sequence $\tens{H}$ equal to 4 and visualize the temporal and spatial attention weights in Figure \ref{temproal_4} left and right, respectively.
The agent's latest observation is $\tens{O_t}$, and its historical observations are $\tens{O_{t-1}}$, $\tens{O_{t-2}}$, and $\tens{O_{t-3}}$.
For temporal attention, although most historical representations focus on the last frame, the focus on other time steps is not 0.
The right one in Figure \ref{temproal_4} more clearly indicates the emphasis on historical information for spatial attention.
Specifically, the latest observation $\tens{O_{t}}$ not only pays attention to itself but also pays attention to the early historical memory $\tens{O_{t-3}}$.
These results demonstrate that although the latest observation is important for policy, historical information also plays an indispensable role.



\subsection{More Case Studies}
\begin{figure*}
  \centering
  \includegraphics[width=1\linewidth]{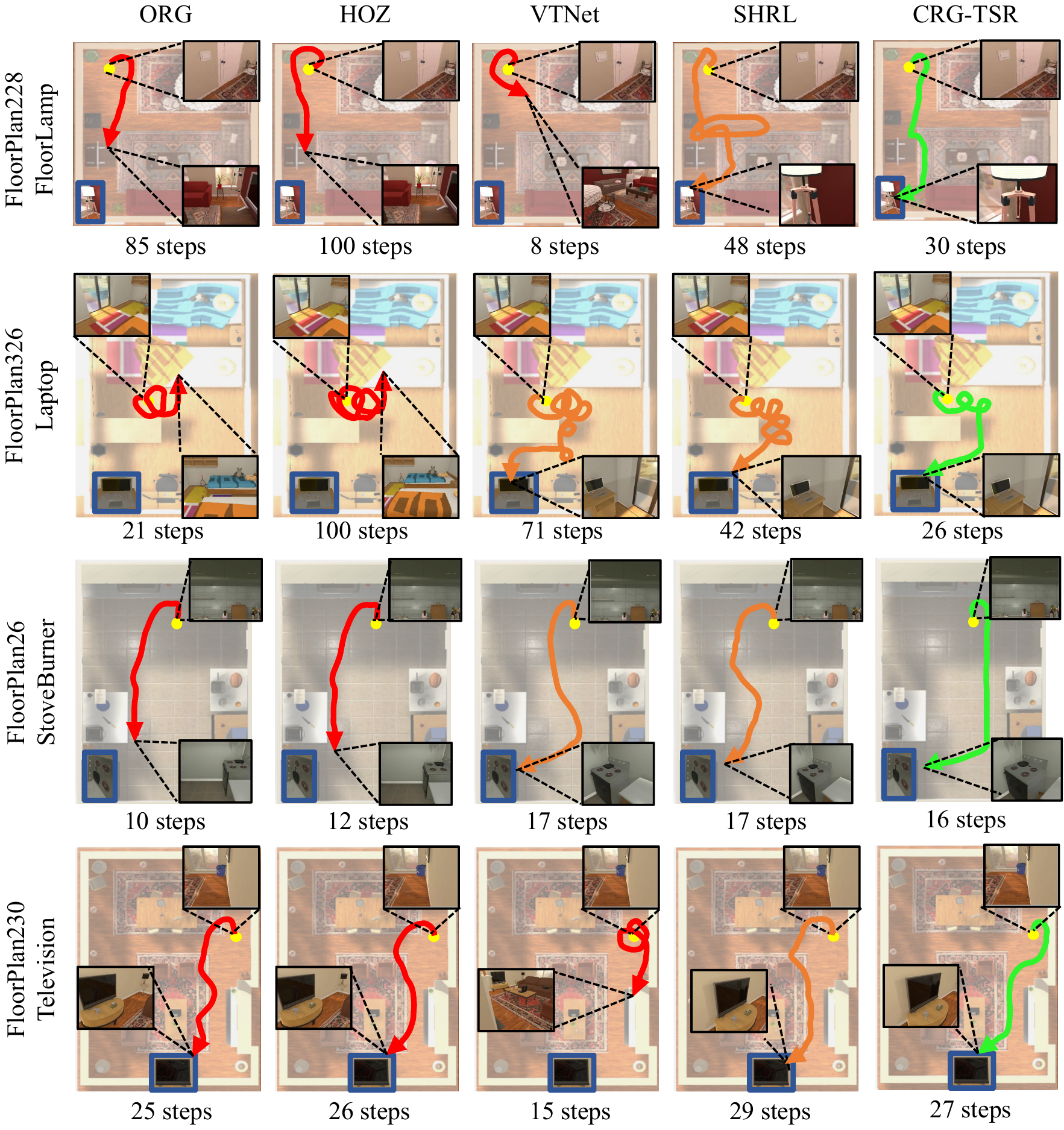}
  \caption{\textbf{Visualization results of different models in the test environment. }
The target object is highlighted with blue bounding boxes, and the starting point is marked with a yellow dot.
  The green line represents the best case, the orange line represents the sub-optimal case, and the red represents the failed case. 
  We show six cases, then label the room number and target category on the left side of each row.
  The episode produced by our agent successfully reaches the target and has the shortest steps. 
In comparison, all other agents often fail due to obstacles, inaccurate distance perception, or taking more steps. 
Our method maintains superiority in more complex environments.
  }
  \label{case_study}
\end{figure*}

We compared the trajectories of six navigation tasks proceeded by four methods, \textit{i.e.}, ORG, HOZ, VTNet, SHRL, and our CRG-TSR method, in Figure \ref{case_study}.
Experimental results show that our model has more effective and efficient navigation trajectories.
In contrast, other methods often take more steps or fail due to the occlusion of obstacles and inaccurate position perception of targets.


\section{Impact of Pre-training.}
\label{sec:Pre-training}
To study the influence of the quality of different pre-training trajectory datasets, we vary the method to generate the observation sequence, \textit{i.e.}, Random, ORG, VTNet, Expert.
The expert policy utilizes the offline map and Dijkstra's algorithm to generate the optimal action. 
Simultaneously, we also report the performance of the method without pre-training.
We compare variants of different methods in Table \ref{pretrain}.
For the training episodes, the results show quantitatively that the pre-training phase reduces the episode from $6M$ to $2M$. 
For the model performance, the high-quality trajectory data can bring improvement, and poor-quality data will have a negative impact, but the effect is insignificant.
Specifically, when utilizing the expert policy to generate the pre-training dataset, the performance of our method is improved from 79\% to 80\% in the formal training.
The random policy trajectory datasets also improve the convergence speed but may converge to a sub-optimal policy.
Those results show that our model is stable for the pre-training stage, regardless of whether it performs or utilizes which pre-training datasets.

\begin{table*}[ht]
\centering
\caption{\textbf{Impacts of different methods to generate pre-training dataset.} 
The symbol g. is an abbreviation of "generate," and w/o means "without." 
We report the  Success Rate (SR), Success Weighted by Path Length (SPL), and their variances in parentheses by repeating experiments five times.
The best performance is achieved using the Expert policy to generate the datasets.}
\resizebox{0.8\textwidth}{!}{
\begin{tabular}{L{3cm}|c|cc|cc}
\hline
\multirow{2}{*}{Method} & \multirow{2}{*}{Episodes} & \multicolumn{2}{c}{ALL} \vline  & \multicolumn{2}{c}{$L_{opt}\geqslant 5$}\\[1pt] \cline{3-6}
&  & \makecell[c]{SR} & \makecell[c]{SPL}   &   \makecell[c]{SR}  & \makecell[c]{SPL}\\[1pt]\hline \hline
g. by VTNet  &2M  &79.4(0.524) &0.468(0.008) &72.3(0.634) &0.450(0.009)    \\
g. by ORG  &2M &79.3(0.694)  &0.451(0.003) &72.5(0.490) &0.448(0.004)    \\
g. by Random  &2M &77.1(0.386)  &0.446(0.004) &69.4(0.732) &0.434(0.004) \\
w/o pre-training &6M  &79.4(0.524)  &0.462(0.002) &72.7(0.282) &0.458(0.002)    \\
\hline 
\textbf{g. by Expert(Ours)} &\textbf{2M}    &\textbf{80.0(0.825)}  &\textbf{0.456(0.005)}  &\textbf{73.6(1.131)}  &\textbf{0.451(0.007)}  \\\hline
\end{tabular}}
\label{pretrain}
\end{table*}
